\documentclass[lettersize,journal]{IEEEtran}
\usepackage{amsmath,amsfonts}
\usepackage{algorithmic}
\usepackage{array}
\usepackage[caption=false,font=normalsize,labelfont=sf,textfont=sf]{subfig}
\usepackage{textcomp}
\usepackage{stfloats}
\usepackage{url}
\usepackage{verbatim}
\usepackage{graphicx}
\usepackage{cite}
\usepackage{threeparttable} 
\usepackage{booktabs}
\usepackage{adjustbox} 
\usepackage{multirow}  
\usepackage{booktabs}  
\usepackage{diagbox}   
\usepackage[linesnumbered,ruled,vlined]{algorithm2e}
\usepackage{amsmath,amssymb,amsfonts}
\usepackage{tabularx} 

\begin{document}

\title{A Sliding-Window-Based Reinforcement Learning for Dynamic Assembly Flow Shop Scheduling with Multi-Product Delivery}

\author{Junhao Qiu, Jianjun Liu, Ting Liu, Rongjie Liao, Zhantao Li, Qingfu Zhang,~\IEEEmembership{Fellow,~IEEE}
 \thanks{This research was supported by the National Natural Science Foundation of China (Grant no: 52375489, 51975129) and the Guangdong Basic and Applied Basic Research Foundation of China (Grant no: 2024A1515011211). \textit{(Corresponding author: Jianjun Liu.)} }
 \thanks{Junhao Qiu is with the School of Electromechanical Engineering, Guangdong University of Technology, Guangzhou, China, and also with the Department of Computer Science, City University of Hong Kong, Hong Kong (Email: junhaoqiu2-c@my.cityu.edu.hk).}
 \thanks{Jianjun Liu, Ting Liu, Rongjie Liao and Zhantao Li are with the School of Electromechanical Engineering, Guangdong University of Technology, Guangzhou, China (Email: jianjun.liu@gdut.edu.cn, 1112401015@mail2.gdut.edu.cn, liaorongjie1@mails.gdut.edu.cn, lzt@gdut.edu.cn).}
\thanks{Qingfu Zhang is with the Department of Computer Science, City University of Hong Kong, Hong Kong (Email: qingfu.zhang@cityu.edu.hk).}
}

\maketitle

\begin{abstract}

Multi-product kitting delivery imposes significant challenges for real-time scheduling in hybrid manufacturing systems that integrate processing and assembly, as dynamic order arrivals simultaneously alter supply dependencies and the set of feasible job-machine assignments. This paper proposes a sliding-window-based reinforcement learning (SWRL) framework for end-to-end online scheduling in the flexible assembly flow shop scheduling problem with complex kitting constraints. The problem is formulated as a heterogeneous graph-based Markov decision process that captures the dual-layer kitting structure and the tail-product bottleneck dynamics that produce a sparse reward landscape. To address the resulting challenges, SWRL integrates a sliding-window filtering mechanism that filters inactive nodes and prioritizes kitting-critical operations, a spatiotemporal graph encoding network that tracks bottleneck shifts across consecutive decision states, and a dynamic action mapping module with a constrained waiting strategy that adapts to the changing action space under variable topologies. Experiments on real-world instances from a home appliance manufacturer demonstrate that SWRL achieves consistent tardiness reductions over classical dispatching rules and existing deep reinforcement learning methods, and exhibits robust performance across varying resource configurations, order loads, and arrival concentrations.

\end{abstract}


\section{Introduction}
\label{introduction}
\IEEEPARstart{T}{he} increasing demand for scenario-based suite orders has driven hybrid manufacturing systems that integrate processing and assembly toward customer-centric production models~\cite{8821409, chen2022unmanned}. Traditionally, coordination between processing and assembly stages has relied on inventory buffers~\cite{Zhao2018Manufacturing}, but dynamic order arrivals render such static buffers ineffective. In such systems, uncertainty in order arrivals repeatedly destabilizes planned schedules and degrades delivery performance. Online scheduling methods that can respond to multi-stage disruptions and track kitting progress across parallel product flows are therefore essential for the dynamic flexible assembly flow shop scheduling problem with multi-product delivery (DFAFSP-MPD).

The flexible assembly flow shop scheduling problem (FAFSP) is an NP-hard extension of the classical flow shop problem, incorporating part kitting assembly under flexible resource constraints~\cite{liu2025recent}. It is widely applied in industries such as home appliances~\cite{liao2025collaborative} and automotive~\cite{Liao2015efficient}. The difficulty of DFAFSP-MPD originates from its dual-layer kitting structure. At the product level, assembly kitting requires all component parts to be completed before final assembly. At the order level, multi-product delivery requires all products within a single order to be completed simultaneously, such that the order lead time is determined by its tail product. Under dynamic arrivals, the identity of this tail product shifts with sequence-dependent setup times and varying processing speeds, producing a moving bottleneck that is difficult to predict. The resulting decision space requires both job sequencing and machine assignment under flexible processing qualifications, where the tardiness signal is sparse across actions.

Centralized static methods, including mixed-integer linear programming~\cite{Wu2021Metaheuristics} and metaheuristics~\cite{li2025dynamic, 9711566}, provide effective look-ahead optimization but are computationally infeasible for real-time decisions at the second or millisecond scale~\cite{lei2023large}. Approximate methods~\cite{Hatami2015Heuristics} offer a practical trade-off for moderate-scale instances. Robust scheduling approaches embed disturbance tolerance through time buffers and scenario analysis~\cite{feng2016robust}, while rescheduling methods repair plans after disruptions~\cite{rahmani2016stable, guo2024multi}. However, their effectiveness degrades under high-frequency or large-scale disturbances, a common condition in online FAFSP environments~\cite{yang2022real, li2025real}. These limitations highlight the need for adaptive, data-driven scheduling policies learned directly from system interactions.

Deep reinforcement learning (DRL) has been widely adopted for dynamic scheduling due to its adaptive decision-making capability~\cite{mnih2015human,wang2021review}. One approach uses DRL to select priority dispatching rules (PDRs) at each decision step, formulated as a Markov decision process (MDP)~\cite{zhang2023counterfactual, 9673698}. \cite{luo2020dynamic} applied deep Q-network (DQN) for dispatching rule selection in FJSP, while~\cite{yang2022real, yang2023real} extended this to distributed FSP settings. \cite{wang2022multi} further improved generalization through multi-objective DQN. PDR-based methods provide a guaranteed performance lower bound and can be extended to complex shop floors~\cite{wang2022independent, qiu2024multi, luo2021real, wang2021adaptive}, but their dependence on fixed rule sets limits attainable performance. An alternative paradigm employs end-to-end DRL solvers with graph representations to directly allocate operations and resources~\cite{liu2022graph, li2025graph, liu2023dynamic}. \cite{liu2022graph} combined GNNs with DRL for JSP, using graph representations for state encoding and proximal policy optimization for scheduling decisions. \cite{song2022flexible} applied a heterogeneous graph neural network to model FJSP states and achieved competitive makespan performance. This approach was further extended to dynamic job arrivals by~\cite{zhang2022dynamic}. \cite{lei2023large} proposed a hierarchical reinforcement learning framework for large-scale dynamic FJSP, using a high-level agent to manage job release and lower-level agents for operation sequencing. However, these end-to-end methods are primarily designed for static graph structures~\cite{lei2022multi} and do not address the challenges specific to DFAFSP-MPD, where the graph topology and action space evolve dynamically due to assembly kitting constraints~\cite{park2026deep}.

Applying existing end-to-end DRL methods to DFAFSP-MPD introduces three technical challenges.
1) \textit{State representation under dual kitting constraints.} Multi-product delivery couples parallel products through a $\max$ operator: each order's tardiness depends only on its tail product. This produces a stepped reward surface where most job-machine pairs cause no immediate change in tardiness, impeding the agent's ability to identify bottleneck-critical actions~\cite{challenge1}. Simultaneously, dynamic order arrivals expand the candidate job set, submerging bottleneck-related signals in noise from irrelevant parallel operations.
2) \textit{Capturing spatiotemporal correlations in dynamic environments.} The tail product bottleneck is not static, but shifts as sequence-dependent setup times and processing speeds alter each product's relative completion progress. Existing methods that encode the scheduling state as a static graph at each decision step fail to capture the temporal dependencies between consecutive states~\cite{lei2023large}, making it difficult for the agent to anticipate where kitting pressure will accumulate next.
3) \textit{Sparse reward under complex constraints.} The dual-layer kitting structure creates a highly constrained action space where only a limited subset of job-machine pairs reduces order tardiness at any given step. Most actions yield zero immediate reward, and each decision step advances only a single job~\cite{qiu2024multi}. The resulting reward signal is both sparse and delayed, providing limited gradient information for policy optimization, a problem that intensifies with problem scale.

To address these challenges, this paper proposes a sliding-window-based reinforcement learning (SWRL) framework. SWRL integrates three corresponding components: a sliding-window-based filtering mechanism that filters inactive nodes to mitigate feature dilution, a spatiotemporal graph encoding network that tracks bottleneck shifts across consecutive decision states, and a dynamic action mapping module with a constrained waiting strategy that adapts to changing action spaces under sparse reward conditions. The contributions of this research are summarized as follows:

\begin{enumerate}
	\item The DFAFSP-MPD is formulated as a mixed-integer linear programming model and a heterogeneous graph-based MDP. The model captures the dual kitting constraints including assembly-level part coupling and order-level product synchronization, as well as the tail-product bottleneck dynamics that create a sparse and shifting reward landscape.

	\item A sliding-window-based reinforcement learning framework, termed SWRL, is proposed for end-to-end online scheduling under variable topologies and action spaces. A sliding-window filtering mechanism dynamically filters inactive job nodes and prioritizes kitting-critical ones as the graph structure evolves. Node features from temporally correlated graph states are extracted by a spatiotemporal graph neural network for state representation. A dynamic action mapping module with a constrained waiting strategy is then adopted to align feasible job-machine pairs under changing action spaces.

	\item Comprehensive experiments are conducted on both generated and real-world instances, including parameter sensitivity analysis, ablation studies, comparative evaluation with state-of-the-art DRL-based scheduling methods, and robustness verification across varying problem scales and arrival patterns. SWRL consistently outperforms existing approaches in tardiness minimization while requiring substantially fewer training episodes. Ablation studies further validate the contribution of each designed component to the overall performance.
\end{enumerate}

The remainder of this paper is organized as follows. Section \ref{PROBLEM} describes the FAFSP and its heterogeneous graph modeling. Section \ref{Method} presents the SWRL framework. Section \ref{Experiment} presents experimental results and section \ref{Conclusion} concludes this work.





\section{Problem Formulation of FAFSP-MPD}
\label{PROBLEM}

\subsection{Problem Description}
DFAFSP-MPD is derived from a real-world assembly and processing workshop of home appliance companies. This processing-assembly hybrid manufacturing system consists of ${{M}_{p}}$ unrelated processing machines and ${{M}_{a}}$ assembly lines. Its problem description is as follows:

1) \textit{Flexible processing capability}: The qualifications and processing efficiency of machines at different stages vary when performing the same job. Switching between job $j_1$ and job $j_2$ of different products requires a setup time of $S{{T}_{{{j}_{1}},{{j}_{2}}}}$.

2) \textit{Assembly kitting constraint}: The products of order $i$ are divided into corresponding processing and assembly jobs $j$ according to their structure. All supporting parts need to be completed before final assembly to ensure the hierarchical coupling constraints of supply.

3) \textit{Multi-product delivery}: Each order $i$ has a specific delivery time $DT_i$, containing at least one product. All products in the order are completely assembled before delivery.

4) \textit{Dynamic arrival of orders}: With orders arriving dynamically, it is necessary to respond quickly to disruptions and adjust the planned schedule, coordinating processing and assembly resources to reduce delays for multi-product orders.

The notations used in this article are summarized as follows:

\begin{tabularx}{\linewidth}{@{}l@{\ }l@{\ }X@{}}
$i$        & $:$ & Order index, $i\in I$ \\
$p$        & $:$ & Product index, $p\in P$ \\
$j$        & $:$ & Job index, $j_1,j_2\in J$ \\
$m$        & $:$ & Machine index, $m\in M$ \\
$Q_{j,m}$  & $:$ & 1 if machine $m$ is qualified to process job $j$, 0 otherwise \\
$P_{j_1,j_2}$ & $:$ & 1 if job $j_1$ is the immediate predecessor of job $j_2$, 0 otherwise \\
$PT_{j,m}$ & $:$ & Processing time of job $j$ on machine $m$ \\
$A_{p,j}$  & $:$ & 1 if job $j$ is an assembly task for product $p$, 0 otherwise \\
$O_{i,p}$  & $:$ & 1 if product $p$ belongs to order $i$, 0 otherwise \\
$DT_i$     & $:$ & Delivery time of order $i$ \\
\end{tabularx}
\begin{tabularx}{\linewidth}{@{}l@{\ }l@{\ }X@{}}
$AT_i$     & $:$ & Arrival time of order $i$ \\
$ST_{j_1,j_2}$ & $:$ & Setup time for switching from job $j_1$ to job $j_2$ \\
$V$        & $:$ & A sufficiently large positive constant \\
$tt_i$     & $:$ & Tardiness of order $i$ \\
$ft_i$     & $:$ & Completion time of order $i$ \\
$ct_j$     & $:$ & Completion time of job $j$ \\
$st_j$     & $:$ & Start time of processing for job $j$ \\
$x_{j_1,j_2,m}$ & $:$ & 1 if job $j_1$ is processed immediately after job $j_2$ on machine $m$, 0 otherwise
\end{tabularx}

The goal is to minimize the total tardiness while satisfying the following assumptions and constraints.

1) This study does not consider transportation.

2) Each machine can process only one job at a time.

3) Once the operation has started, no interruption is allowed.

4) A job can only proceed to the assembly stage after all its components have been fully completed.

The objective is to minimize the total tardiness of orders, as defined in~\ref{obj}.
\begin{equation}
\label{obj}
\min \mathop{\sum }_{i=1}^{I}\left( {tt}_{i} \right)
\end{equation}

\textit{Order delays for tail products}: The tardiness of orders $i$ varies drastically, with the final delay depending on the completion time of the last product, $ft_i = \max \limits_{\forall p,j}(ct_j \times A_{p,j} \times O_{i,p})$. Constraints can be represented as~\ref{con01} and~\ref{con02}.
\begin{equation}
\label{con01}
\begin{aligned}
ft_i \ge ct_j +V \times \left( A_{p,j} + O_{i,p}-2 \right), \\
\forall i \in I,p\in P,j \in J
\end{aligned}
\end{equation}
\begin{equation}
\label{con02}
tt_i\ge ft_{i}-DT_i,\forall i \in I
\end{equation}

\textit{Flexibility and hierarchical coupling constraint}: Flexible processing capabilities are represented as~\ref{con03} by qualification matrix, and hierarchical coupling constraints ensure that the order of processing paths is satisfied as~\ref{con04}.
\begin{equation}
\label{con03}
{x}_{j_1,j_2,m}\le {{Q}_{j_2,m}},\forall j_1\in [0]\cup J,j_2 \in J,m\in M
\end{equation}
\begin{equation}
\label{con04}
\begin{aligned}
ct_{j_2}\ge \sum\limits_{m=1}^{M}{\sum\limits_{j_3=0}^{J}{\left( x_{j_3,j_2,m} \times \left( PT_{j_2,m} + ST_{j_3,j_2} \right) \right)}}\\
+ P_{j_1,j_2}\times ct_{j_1},
\forall j_1,j_2 \in J
\end{aligned}
\end{equation}

\textit{Flow and sequence restrictions}: Define the workflow and sequence constraints for jobs on the machine as~\ref{con05} and~\ref{con06}. Note that $j=0$ represents a virtual order, while $ct_j=0$.
\begin{equation}
\label{con05}
\sum_{m=1}^{M} \sum_{j_1 = 0}^{J} x_{j_1, j_2, m} = 1, \forall j_2 \in J
\end{equation}
\begin{equation}
\label{con06}
\sum_{j_1=0}^{J} x_{j_1, j_2, m} - \sum_{j_1=0}^{J} x_{j_2, j_1, m} = 0, \forall j_2 \in J, m \in M
\end{equation}

\textit{Time window limit}: The start and end times of the job are constrained as~\ref{con07}-\ref{con09} to ensure compliance with processing time and arrival time limits.
\begin{equation}
\label{con07}
\begin{aligned}
st_j\ge \left( A_{p,j} + O_{i,p}-2 \right)\times V+AT_i, \\
\forall i \in I,p\in P,j \in J
\end{aligned}
\end{equation}
\begin{equation}
\label{con08}
\begin{aligned}
st_{j_2}\le ct_{j_2}-PT_{j_2,m}+V\times \left( 1-\sum\limits_{{j_1}=0}^{J}{x_{j_1,j_2,m}} \right), \\
\forall j_2 \in J,m \in M
\end{aligned}
\end{equation}
\begin{equation}
\label{con09}
\begin{aligned}
ct_{j_2}\ge  ct_{j_1} \times PT_{j_2,m}+ST_{j_1,j_2}+V\times \left( x_{j_1,j_2,m} -1 \right),\\
\forall j_1 \in [0]\cup J, j_2 \in J,m \in M
\end{aligned}
\end{equation}

The $tt_i, ft_i, ct_j, st_j\ge 0, \forall i\in I,j \in J$ specify the feasible range of decision variables.

\begin{figure*}[h]
    \centering
    \includegraphics[width=1\textwidth]{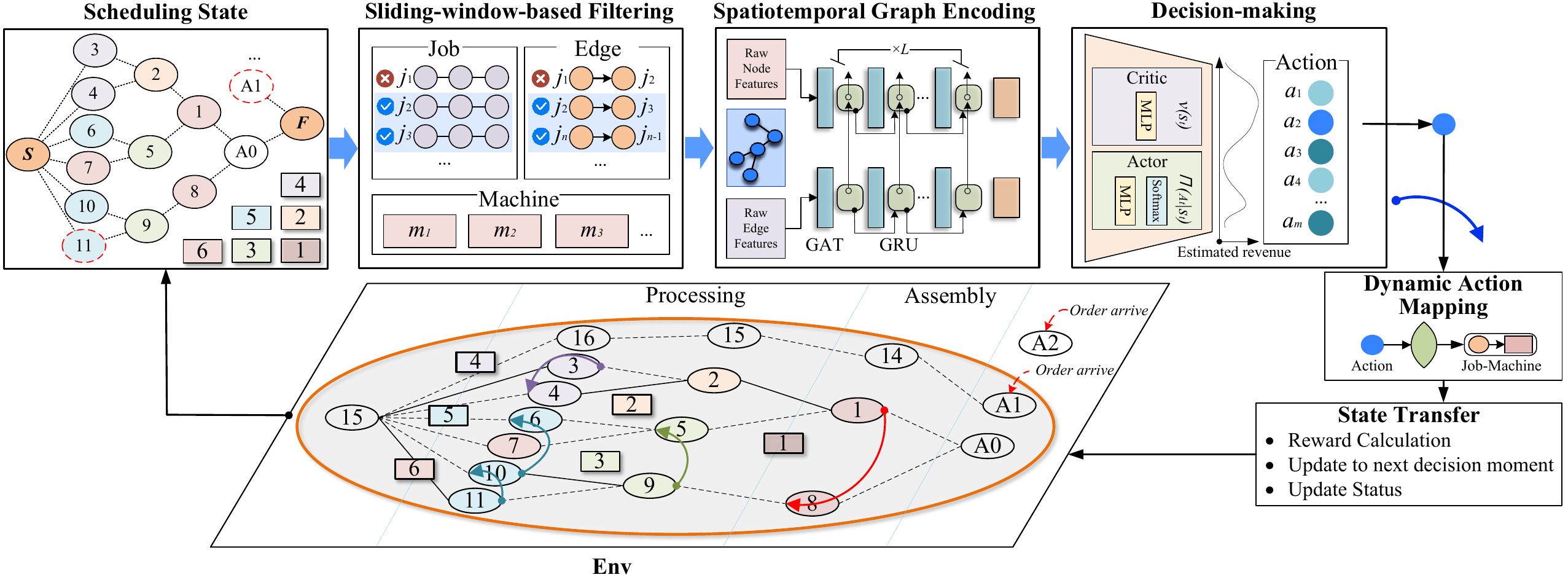}
    \caption{End-to-end online scheduling framework of the proposed SWRL, integrating three modules: sliding-window-based filtering mechanism, spatiotemporal graph encoding, and dynamic action mapping.}
    \label{Fig3}
\end{figure*}

\subsection{Complexity analysis}

The complexity of DFAFSP-MPD arises from the coupling of job sequencing, machine assignment, and hierarchical kitting constraints. Let $N = \sum_{i=1}^{I} \sum_{p=1}^{|P_i|} |J_{i,p}|$ denote the total number of jobs. Since the job sequence within each product is fixed~\ref{con04}, the effective number of sequencing permutations is $Y = N! / \prod_{i,p} (|J_{i,p}|!)$, accounting for repeated operations within each product. For machine assignment, each job $j$ can be processed on any qualified machine $m \in M_j$, yielding $Z = \prod_{j} |M_j|$ possible assignments.

The overall search space $Y \times Z$ is compounded by the tail-product bottleneck dynamics. While the tardiness of each order is determined by the completion time of its last finishing product ($ft_i = \max_{\forall p,j} ct_j \cdot A_{p,j} \cdot O_{i,p}$), the identity of this tail product is not fixed. Sequence-dependent setup times $ST_{j_1,j_2}$ and varying processing speeds $PT_{j,m}$ cause the relative completion progress across products to shift over time, meaning that any job has the potential to become the critical determinant of its order's delivery performance. This creates a decision space where the effective bottleneck moves dynamically across the production horizon.

\subsection{Heterogeneous Graph}
In traditional JSP, multiple directed arcs are commonly used to form start-to-end paths representing the processing sequence of different jobs. Accurately characterizing the job-machine arc features introduced by flexible processing constraints is challenging, as it can significantly increase graph density. In this study, FAFSP is formulated as a structure of heterogeneous graph $\mathcal{H}=\left( \mathcal{N},\mathcal{M},\mathcal{C},\mathcal{E} \right)$, $\mathcal{N}$ is the set of job nodes, and it still adopts the directed acyclic graph to describe the topology of job nodes, $\mathcal{N}=\left\{ {{\mathcal{N}}_{j}}|\forall j\right\}\cup \left\{ Start,End \right\}$. $\mathcal{C}$  is the set of connected arcs of jobs, which describes the processing sequence of all jobs in terms of directed connected arcs. $\mathcal{E}$ is the arc set of job-machine pairs introduced by considering flexible processing characteristics, where ${{\mathcal{E}}_{n_1,n_2}}\in \mathcal{E}$ is an undirected arc connecting different jobs and compatible machine nodes $m$. 
Select a job-machine arc for each node and determine its direction to construct the scheduling scheme for FAFSP.


\section{Method}\label{Method}

This section presents the SWRL framework for online scheduling under variable topologies and action spaces, as illustrated in Fig.~\ref{Fig3}. Solving online FAFSP requires adaptively selecting job-machine pairs at each decision step. The framework comprises three modules that operate sequentially at each step. First, the sliding-window filtering mechanism (Section~\ref{SlidingWindow}) filters inactive job nodes and maintains a fixed-size set of kitting-critical candidate nodes. Second, the spatiotemporal graph encoder (Section~\ref{SpatioTemporal}) extracts features from the filtered graph state, capturing both spatial dependencies and temporal evolution across consecutive decision steps. Third, the dynamic action mapping module (Section~\ref{Action_Transcription}) aligns the policy output to feasible job-machine pairs and executes the selected action. The MDP formulation defines the state, action, reward, and transition dynamics.

\subsection{MDP Formulation}\label{MDP}
\subsubsection{State}
The state $s_t$ at decision step $t$ is represented as a heterogeneous graph $\mathcal{H}_t = (\mathcal{T}_t, \mathcal{M}_t, \mathcal{C}_t, \mathcal{E}_t)$, where $\mathcal{T}_t$ and $\mathcal{M}_t$ are the job and machine node feature sets, and $\mathcal{C}_t$ and $\mathcal{E}_t$ are the arc sets representing job dependencies and job-machine compatibility, respectively. The node features are summarized in Table~\ref{tab02}.

\begin{table}[t]
\caption{Node features}\label{tab02}
\begin{adjustbox}{center}
\centering
\resizebox{1\columnwidth}{!}{
\begin{tabular}{cll}
\toprule
Index & Expression & Description \\
\midrule
$\mathcal{N}_{t,1}$ & $\operatorname{avg}(P_{j,m})$ & Average processing time \\
$\mathcal{N}_{t,2}$ & $\min(T_t, st_j) - AT_t$ & Job waiting time \\
$\mathcal{N}_{t,3}$ & $M_{j,t}$ & Number of compatible machines \\
$\mathcal{N}_{t,4}$ & $\max(ct_j - DT_i, 0)$ & Job tardiness at step $t$ \\
$\mathcal{N}_{t,5}$ & $\{0,1\}$ & Processability flag \\
$\mathcal{M}_{t,1}$ & $WT_{t,m}$ & Machine cumulative idle time \\
$\mathcal{M}_{t,2}$ & $EC_{t,m}$ & Machine setup change count \\
$\mathcal{M}_{t,3}$ & $FJ_{t,m}$ & Completed job count \\
$\mathcal{M}_{t,4}$ & $CJ_{t,m}$ & Available jobs count \\
$\mathcal{M}_{t,5}$ & $\{0,1\}$ & Machine availability \\
\bottomrule
\end{tabular}}
\end{adjustbox}
\end{table}

The arc set $\mathcal{C}_t$ encodes the immediate supply relationships between jobs, where $f_{\mathcal{C}_{j_1,j_2}} = 1$ if the processing condition between $j_1$ and $j_2$ is satisfied at step $t$, and 0 otherwise. The arc set $\mathcal{E}_t$ encodes job-machine pairs, with $f_{\mathcal{E}_{j,m}}$ storing the processing time of job $j$ on machine $m$. Both arc sets are recomputed at each step to reflect changes caused by job completions and new arrivals.

\subsubsection{Action}
The actions $a_t \in A_t$ at step $t$ are an arc set consisting feasible job-machine pairs $(j,m)$ and a waiting action $a_{wait}$. Waiting refers to the decision not to execute any action $(j,m)$ at $t$, with idle machine resources remaining in a state of readiness until the next event triggers a decision~\cite{wang2022independent}.
All the immediately preceding jobs of job $j \in J$ have been completed and satisfy the qualification requirements of machine $m$ while $m \in M$ is in an idle state.
Also, $A_t$ is related to the decision step and needs to be dynamically processed based on the arrival time and supply relationship. Meanwhile, since there are qualification and stage differences among different machines, then $J_m < M$, so $|A_t| < j \times m$. Applying the wait action enables the agent to refrain from executing $(j,m)$ pairs, reserving processing capacity for urgent orders and expanding the agent's pre-decision capabilities. To avoid the dilemma of continuous waiting, a waiting action restriction strategy that considers topology-variable is proposed (Section \ref{Action_Transcription}).

\subsubsection{Reward}
Reward function is structured based on delayed performance and penalties derived from balancing long-term and short-term perspectives. The $ct_{i,t}$ is last processing time of order $i$ at decision step $t$, and $\Delta ct_{i,t}$ denotes the additional delay at the current step at decision step $t$, calculated as:
\begin{equation}
\Delta ct_{i,t} = \max(ct_{i,t} - DT_i, 0) - \max(ct_{i,t-1} - DT_i, 0).
\end{equation}

The $\delta$ is the time step penalty coefficient, satisfying $\delta > 0$. $T_t$ denotes the time of the current decision step, $T_{t-1}$ denotes the time of the previous decision step, and $T_t - T_{t-1}$ denotes the time interval between $t$-$1$ and $t$. The reward is calculated:
\begin{equation}
r_t = -\sum_{i \in I^{\text{all}}_{t}} \Delta ct_{i,t} - \delta (T_t - T_{t-1}).
\end{equation}

Since the tardiness of all orders is $T = \sum_{i=1}^I (tt_i)$, the absolute value of the total reward is equal to this quantity. This implies that the reward function effectively distributes the total order tardiness across the decision-making process. The coefficient $\delta$ introduces a secondary penalty on the elapsed time between decisions, which implicitly discourages unnecessarily long processing sequences. The sensitivity of this trade-off between tardiness and throughput is examined in the orthogonal experiment (Section~\ref{Parameter}).

\subsubsection{Transfer}
After action $a_t$ is executed, the simulation advances to the next decision step. Each decision step is triggered by the earliest event in which either a machine completes its current job and becomes idle, or a new order arrives. At that point, the environment state is updated: completed job nodes and their incident edges are removed from the heterogeneous graph, new nodes from arriving orders are inserted, and the edge features are recomputed to reflect the updated processing conditions. This transition produces a new state $s_{t+1}$ with an updated action set $A_{t+1}$.

\begin{figure}[h]
    \centering  
    \includegraphics[width=1\columnwidth]{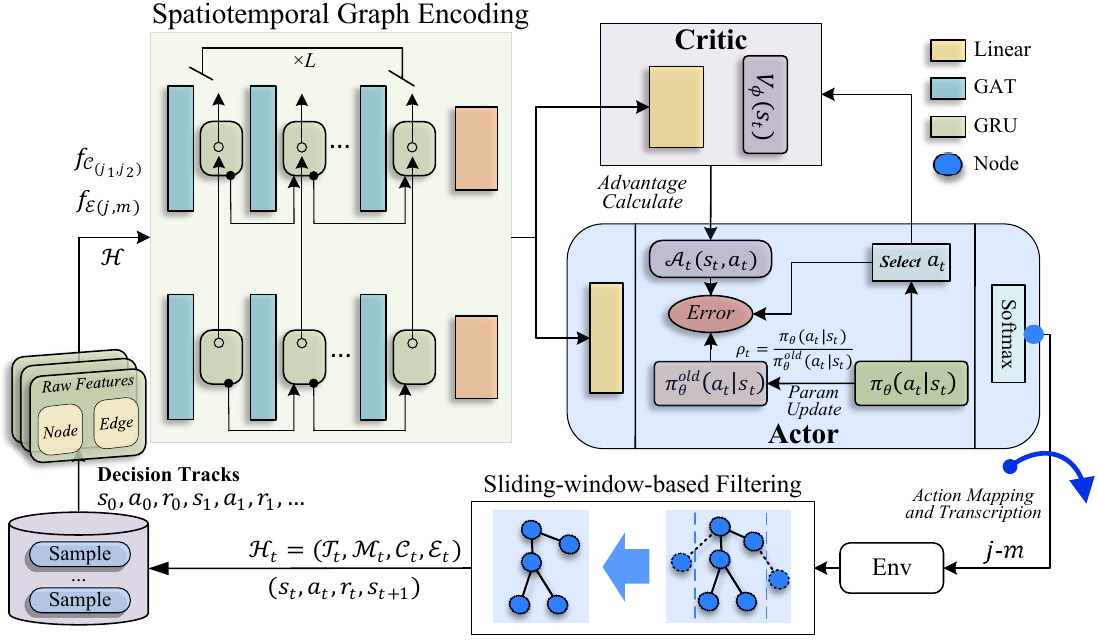}
    \caption{Construction of end-to-end agent variable topology inputs. It supports the normalization of variable inputs and extracts spatiotemporal features from decision trajectories.}
    \label{Fig4}
\end{figure}

\subsection{Agent Architecture}\label{Agent_Construction}

The SWRL agent architecture is illustrated in Fig.~\ref{Fig4}. The filtered graph from the sliding-window filtering mechanism is processed by the spatiotemporal encoder, and the resulting embedding is passed to the policy network for action selection.

\subsubsection{Sliding-Window Filtering Mechanism}\label{SlidingWindow}

As orders arrive dynamically, the total number of job nodes in the heterogeneous graph grows monotonically. Encoding all nodes equally dilutes the features of kitting-critical nodes among a large number of irrelevant parallel operations, impeding the agent's ability to identify bottleneck jobs. The sliding-window filtering mechanism addresses this by maintaining a fixed-size active node set $N_t^w$ (the window).

\begin{figure}[h]
    \centering
    \includegraphics[width=0.9\columnwidth]{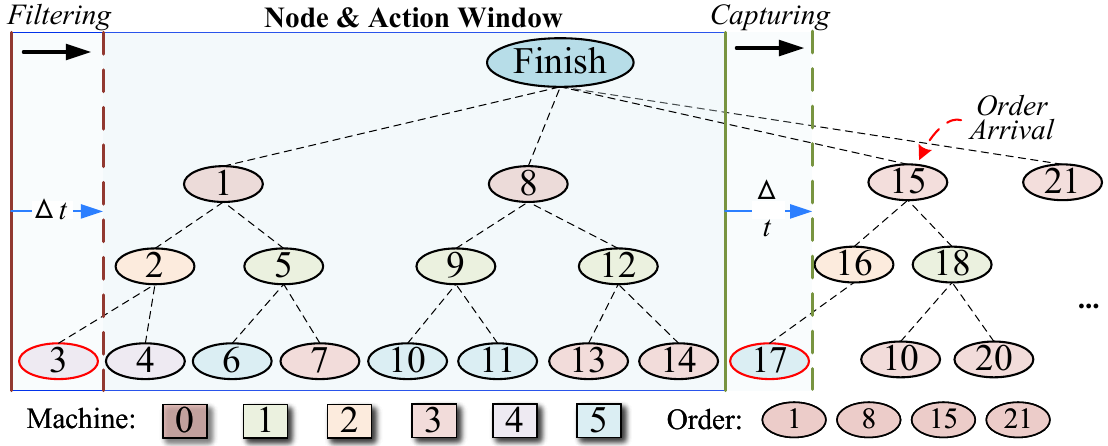}
    \caption{Dynamic filtering and node capture process. At time $\Delta t$, the window filters out completed nodes and captures newly available nodes based on in-degree priority.}
    \label{Fig5}
\end{figure}

As illustrated in Fig.~\ref{Fig5}, each job node becomes active when its predecessors are completed and is removed from the window once finished. The vacated slot is reassigned to a pending node with the highest in-degree $ID_n$ measured within the conjunctive arc set $\mathcal{C}_t$.

When a node exits the window, its associated job-machine pairs are removed from the action space, and the newly admitted node's pairs are inserted. This filtering reduces the action space dimension from the full set of all candidate pairs to a focused set of currently relevant decisions. 

The complete window update and action mapping process is given in Algorithm~\ref{algo1}. The window update proceeds in two phases. First, all finished nodes are identified and their job-machine pairs are removed from the action space (lines 4-14). A finished node is only evicted after all its predecessor jobs have been completed, ensuring that no kitting-critical intermediate state is discarded prematurely. Second, each vacated slot in the eviction phase is filled by a pending node with the minimum in-degree among all candidates outside the current window (lines 15-24). The in-degree is measured within the conjunctive arc set $\mathcal{C}_t$, where lower in-degree indicates that fewer prerequisites remain before the job becomes schedulable. The condition on line 21 checks that the candidate node brings enough feasible job-machine pairs to fill the available action slots; if it does not, the slot is left temporarily empty and will be filled at the next decision step when more nodes become available.

\subsubsection{Spatiotemporal Graph Encoding}\label{SpatioTemporal}

The sliding-window filtering mechanism produces a filtered subgraph that captures the current spatial structure of active jobs and machines. However, the tail product bottleneck shifts over time as setup times and processing speeds alter each order's relative completion progress. A static graph encoding that considers only the current snapshot does not capture this temporal evolution.

To address this, the filtered node and edge features are processed through a spatiotemporal encoding network composed of $L$ stacked layers, each containing a graph attention (GAT) module for spatial aggregation followed by a gated recurrent unit (GRU) for temporal fusion.

The node features $\mathcal{N}_{n,t}$ at step $t$ are first projected into a unified dimension. For any two connected nodes, the attention coefficient is computed as:

\begin{equation}
\label{eq03}
\beta_{n_1,n_2} = \text{LeakyReLU}\left( \eta^T  \left[ W_f \mathcal{N}_{n_1,t}, W_f \mathcal{N}_{n_2,t} \right] \right),
\end{equation}

where $\eta$ is a learnable vector and $W_f$ is a projection matrix. The raw attention scores are then gated by the edge features $f_{(n,n')}$ (which is $f_{\mathcal{C}}$ for job-job edges or $f_{\mathcal{E}}$ for job-machine edges) and normalized via softmax across each node's neighborhood. The normalized coefficient $\alpha_{n,n'} = \mathrm{softmax}(\beta_{n,n'} \cdot f_{(n,n')})$ allows the network to assign different importance to different neighbors depending on both their features and the type of relation (supply dependency or machine compatibility).

\begin{algorithm}[t] \SetKwData{Left}{left}\SetKwData{This}{this}\SetKwData{Up}{up} \SetKwFunction{Union}{Union}\SetKwFunction{FindCompress}{FindCompress} \SetKwInOut{Input}{Input}\SetKwInOut{Output}{output}

	\Input{Action space $A_{t}^{JM}$, set of invalid action indexes $A_{t}^{empty}$ at $t$, set of ($j$,$m$) pairs $JM_{n}$ and predecessor node set of node $n$ is $N_{n}^{s}$.}
	\Output{Action space $A_{t}$ and state ${\mathcal{H}}_{t}$ at step $t$.}
	 \BlankLine

	 \textnormal{Get node space $N_{t}^{w}$ in the window, set of completed job nodes $N_{t}^{nf}$ and filtered nodes $N_{t}^{d}$ at step $t$}\;
	 \ForEach{${n}_{finish}\in N_{t}^{nf}$}{

        \textnormal{Get the Predecessor node set $N_{n_{finish}}^{s}$}\;

        \If{\textnormal{All jobs in $N_{{{n}_{finish}}}^{s}$ have been completed}}{
            \textnormal{Add $n_{finish}$ to $N_{t}^{d}$ and its corresponding index $n^{i}$ in $N_{t}^{nf}$ to $\mathcal{T}_{t}^{w,e}$}\;

            \textnormal{Remove the invalid actions contained in the set $JM_{n_{finish}}$ from $A_{t}^{JM}$}\;

            \textnormal{Add all removed invalid action indexes at $A_{t}^{empty}$ and $n_{finish}$ in $N_{t}^{nf}$}\;
        }
    }
    \If{\textnormal{set of invalidated node indices} $\mathcal{T}_{t}^{w,e}\ne \varnothing $}{
        \ForEach{\textnormal{Indice} $n x^{i} \in \mathcal{T}_{t}^{w,e}$}
        {
	 	\textnormal{Find node $n_{next}$ and the corresponding $JM_{n_{next}}$ based on the minimum $ID_{n}$}\;

            \If{$len(JM_{n_{next}}) \ge len(A_{t}^{empty})$}{

                \textnormal{$\mathcal{T}_{t}^{w}[n x^{i}] \leftarrow n_{next}$}\;

                \textnormal{Insert the $j$-$m$ pair in $JM_{n_{next}}$ into $A_{t}^{JM}$ according to $A_{t}^{empty}$}\;

                \textnormal{Remove $n x^{i}$ from $\mathcal{T}_{t}^{w,e}$ and the index of the inserted action from $A_{t}^{empty}$}\;
            }
        }
    }
    \textnormal{${{A}_{t}}\leftarrow A_{t}^{JM}$ and update ${\mathcal{H}}_{t}$}\;
    \caption{Sliding-window-based action mapping .}
    \label{algo1}
\end{algorithm}

At layer $l$, the GAT module produces node embeddings $h_{n,t}^{(l)}$ by attending over neighbors. These embeddings are then passed to a GRU that merges them with the hidden state $h_{n,t-1}^{(l)}$ from the previous decision step, following the standard GRU formulation. This temporal connection allows the network to track how each node's completion progress evolves across consecutive states.

After $L$ layers, a global embedding for decision-making is obtained by pooling the final node representations:

\begin{equation}
\label{eq04}
h^{\text{global}}_t = \text{concat}\left( \frac{1}{|B_j|} \sum_{j \in B_j} h_{j,t}^{\prime(L)}, \frac{1}{|B_h|} \sum_{n \in B_h} h_{n,t}^{\prime(L)} \right),
\end{equation}

where $B_j$ is the set of job nodes and $B_h = \mathcal{N}_j \cup \mathcal{N}_m$ is the set of all nodes. The resulting embedding encodes both the spatial structure of the current graph and the temporal trends accumulated over past decision steps.

\subsubsection{Action Mapping and Transcription}\label{Action_Transcription}

\begin{figure}[h]
    \centering
    \includegraphics[width=1\columnwidth]{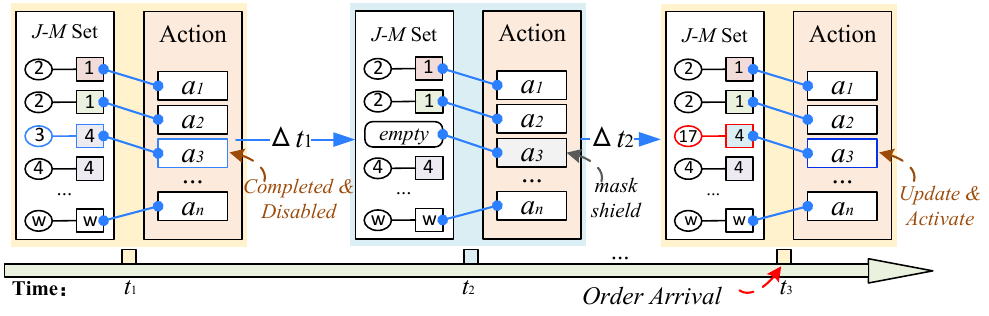}
    \caption{Dynamic mapping between action indices and job-machine pairs. Action indices are reassigned as jobs complete and new nodes enter the window.}
    \label{FigAction}
\end{figure}

The action space at step $t$ consists of the job-machine pairs within the current window plus a waiting action $a_0$. As the sliding window updates, the set of feasible pairs changes: pairs corresponding to completed jobs are removed, and new pairs become available when new nodes enter the window. To maintain a fixed policy output dimension despite this changing set, a mapping table $\mathcal{M}_t: k \rightarrow (j,m)$ is maintained. When a job-machine pair is removed, its index $k$ is immediately reassigned to a newly available pair, a process referred to as index recycling, as illustrated in Fig.~\ref{FigAction}.

The waiting action $a_0$ allows the agent to defer a decision, reserving capacity for future urgent arrivals. To prevent indefinite waiting, the number of consecutive waiting actions is constrained by $a_0^w \leq j_t^{\text{out}}$, where $j_t^{\text{out}}$ is the number of pending nodes outside the current window. This ensures that after at most $j_t^{\text{out}}$ waiting steps, a new node must enter the window and force a processing decision.

Given the global embedding $h_t^{\text{global}}$, the policy network outputs a mean vector $\nu_t$ and a standard deviation vector $\sigma_t$. A binary mask $A_{t,k}^M \in \{0,1\}$ is applied to restrict actions to feasible pairs. The masked log probability used for policy gradient computation is:

\begin{equation}
\label{eq05}
\pi_\theta^{\log} = \sum_{k\in A_t} A_{t,k}^M \left(
\frac{(a_{t,k} - \nu_{t,k})^2}{\sigma_{t,k}^2} +
\log(2\pi \sigma_{t,k}^2)
\right).
\end{equation}

The mask is constructed according to the current mapping table and the waiting constraint, as detailed in Algorithm~\ref{algo2}. Only actions with $A_{t,k}^M = 1$ contribute to the gradient, ensuring that the policy learns only from feasible decisions.

\IncMargin{1em}
\begin{algorithm}[t] \SetKwData{Left}{left}\SetKwData{This}{this}\SetKwData{Up}{up} \SetKwFunction{Union}{Union}\SetKwFunction{FindCompress}{FindCompress} \SetKwInOut{Input}{Input}\SetKwInOut{Output}{output}
	\Input{Nodes spaces $N_t^w$ in the window, waiting action $a_0$, number of uncaptured nodes outside the window $j_t^{out}$, policy network $\pi_\theta$.}
	\Output{Action $a_t$ and the corresponding $j$-$m$ pair}
	 \BlankLine
	 \textnormal{Get action space $A_t$, current number of waiting action $a^w_0$, and initialize the mask $A_{t,a}^{M}$ corresponding to all actions to 0}\;
	 \ForEach{$a\in A_t$}{
        \If{$a = a_0$ \textnormal{and} $N_t^w \ne \varnothing$ \textnormal{and} $a^w_0 \le j_t^{out}$}{
             $A_{t,a_0}^M \leftarrow 1$\;
        }
        \ElseIf{\text{Action} $a$ \text{is valid}}{
            Get the $j$-$m$ pair represented by action $a$\;
            \If{$j$ \textnormal{and} $m$ \textnormal{meet start requirements}}{
                 $A_{t,a}^M \leftarrow 1$\;
            }
        }
    }
    \textnormal{Execution action $a_t$ is obtained from $\max \left(\pi_\theta(a_t | s_t) \right)$ and the mask $A_{t,a}^M$}\;
    \If{$a = a_0$}{
        ${a^w_0}+=1$
    }
    \Else{
    \textnormal{Find corresponding $j$ and $m$ for $a$ from $A_ {t} ^ {JM}$}\;
    }
    \caption{Action transcription mechanism.}
    \label{algo2}
\end{algorithm}
\DecMargin{1em}

The overall scheduling process proceeds as follows. At each decision step $t$, the current heterogeneous graph $\mathcal{H}_t$ is first processed by the sliding-window filtering mechanism to produce a filtered subgraph and action set. The spatiotemporal encoder then computes the global embedding $h_t^{\text{global}}$ from the filtered graph, and the policy network outputs an action distribution. After applying the mask from Algorithm~\ref{algo2}, the action with the highest probability is selected and executed, which assigns a specific job $j$ to a machine $m$ or chooses the waiting action. The simulation then advances to the next event-driven decision step, and the transition $(s_t, a_t, r_t, s_{t+1})$ is stored in the replay buffer for training. This process repeats until all orders are completed.

The policy is trained using proximal policy optimization with generalized advantage estimation. The total objective combines the clipped policy loss (computed over unmasked actions) with a value estimation error term. Network hyperparameters and training details are provided in Section~\ref{Parameter}.

\section{EXPERIMENTAL STUDY}
\label{Experiment}
This section describes the experimental setup and presents results on real-world and extended instances to validate the SWRL framework. A series of experiments are conducted: parameter sensitivity analysis, training evaluation, ablation study, comparative evaluation, and robustness verification.

\subsection{Experimental Setup}
\subsubsection{Evaluation Instances}
The real-world instances are constructed from annual order and product data of a home appliance manufacturer in China, encompassing diverse resource configurations and order demands across multiple factories and time periods. To compensate for the lack of standardized benchmarks for online FAFSP, additional test instances with varying scales are generated following established practices in dynamic scheduling research.

The resource allocation is represented by the machine configuration \( m_1 \times m_2 \). Ten generic product structures ranging from simple to complex are selected from actual production data. The number of products \(q\) per order follows a geometric distribution \(P(q=k) = (1/2)^k\) with \(k\in[1,2,\ldots]\). The processing speed \(v\) of assembly lines follows a normal distribution \(v \sim \mathcal{N}(8, 2)\), and the average assembly speed is four-fifths of the processing stage speed.

\begin{equation}
\begin{aligned}
\label{eq24}
L = \max \Biggl\{
    &\max_{m \in PM} \biggl\{ \sum_{j \in J_p} p_{j,m} \biggr\} + \min_{m \in AM} \biggl\{ \sum_{j \in J_a} p_{j,m} \biggr\}, \\
    &\max_{m \in AM} \biggl\{ \sum_{j \in J_a} p_{j,m} \biggr\}
\Biggr\}
\end{aligned}
\end{equation}
\begin{equation}
\label{eq25}
    Rm_i = \left( DT_i - \max_{m \in M} (p_{j,m} \times B_{i,j}) \right) \times d, \quad \forall i \in I
\end{equation}

Following prior studies \cite{basir2018bi,allahverdi2016minimizing,qiu2024multi}, order delivery times \(DT_i\) are determined by a uniform distribution over \(\left[ L(1-T-R/2), L(1-T+R/2) \right]\), where \(L\) is the lower bound of completion time from (\ref{eq24}) and parameters \(T,R \in [0,1]\). Order arrival times satisfy \(RT_i \in [0, Rm_i]\) sampled from a normal distribution. By varying \(d \in \{0.3,0.5,0.7,0.9\}\), instances with different arrival frequencies are generated to test robustness against disturbances. The resulting experimental conditions are summarized in Table~\ref{tab06}.

\subsubsection{Implementation Details}
All experiments are run on the same hardware platform: an Intel Core i9-13900K CPU and an NVIDIA RTX 3060 GPU (12 GB). The framework is implemented in Python (3.8+) using NumPy (1.24.3) and PyTorch (2.0.1+). The spatiotemporal encoding network uses \(L = 3\) stacked GAT-GRU layers with a hidden dimension of 64 and 4 attention heads. The policy and value networks each consist of two fully connected layers with 128 hidden units. The initial learning rate is set to \(3 \times 10^{-4}\) and decays over the course of training. All DRL-based methods are trained for 2000 episodes per instance set and evaluated on 20 independently generated instances~\cite{luo2020dynamic, song2022flexible, wang2023flexible, qiu2024multi}. All methods share identical data splits and evaluation protocols to ensure fair comparison. 

\subsection{Parameter Sensitivity Analysis}\label{Parameter}
The SWRL framework introduces several hyperparameters that affect convergence and scheduling performance. A Taguchi orthogonal experiment is conducted to analyze the sensitivity of five key factors: the PPO clipping threshold \(\xi\), the time step penalty coefficient \(\delta\), the reward discount factor \(\gamma\), the number of policy replay epochs \(K_{\text{epochs}}\), and the experience replay ratio \(\mu\). Each parameter is assigned four levels, as shown in Table~\ref{tab03}.

\begin{table}[h]
\caption{Orthogonal experiment results for parameter sensitivity}\label{tab03}
\begin{adjustbox}{center}
\centering
\resizebox{1\columnwidth}{!}{
\begin{threeparttable}
\begin{tabular}{l|ll|ll|ll|ll|ll}
\toprule
\multicolumn{1}{c|}{\multirow{2}{*}{Level}} & \multicolumn{2}{c}{$\xi $} & \multicolumn{2}{c}{$\delta $} & \multicolumn{2}{c}{$\gamma $} & \multicolumn{2}{c}{${K}_{\text{epo}}$} & \multicolumn{2}{c}{$\mu $} \\ \cmidrule(l){2-11}
\multicolumn{1}{c|}{} & \multicolumn{1}{c}{Value} & \multicolumn{1}{c|}{$T_{\text{avg}}$} & \multicolumn{1}{c}{Value} & \multicolumn{1}{c|}{$T_{\text{avg}}$} & \multicolumn{1}{c}{Value} & \multicolumn{1}{c|}{$T_{\text{avg}}$} & \multicolumn{1}{c}{Value} & \multicolumn{1}{c|}{$T_{\text{avg}}$} & \multicolumn{1}{c}{Value} & \multicolumn{1}{c|}{$T_{\text{avg}}$} \\ \midrule
1 & 0.1 & 4183 & 0.1 & 4234 & 0.9 & 4312 & \textbf{4} & \textbf{4148} & \textbf{0.3} & \textbf{4100} \\
2 & 0.2 & 4306 & \textbf{0.3} & \textbf{4164} & 0.94 & 4232 & 6 & 4295 & 0.5 & 4190 \\
3 & \textbf{0.25} & \textbf{4163} & 0.5 & 4279 & \textbf{0.97} & \textbf{4121} & 8 & 4236 & 0.7 & 4294 \\
4 & 0.3 & 4204 & 0.7 & 4180 & 0.99 & 4191 & 10 & 4177 & 0.9 & 4272 \\ \bottomrule
\end{tabular}
\begin{tablenotes}
\footnotesize
\item Note: Each parameter is assigned 4 levels with average tardiness \(T_{\text{avg}}\) as the response variable. The optimal level for each parameter is shown in bold.
\end{tablenotes}
\end{threeparttable}
}
\end{adjustbox}
\end{table}

Rank-order analysis of the response ranges indicates that the relative influence on tardiness follows the order \(\mu > \gamma > K_{\text{epochs}} > \xi > \delta\). The experience replay ratio and the discount factor are the most influential parameters, while the time step penalty has the smallest impact. Based on these results, the optimal parameter combination \(\xi = 0.25\), \(\delta = 0.3\), \(\gamma = 0.97\), \(K_{\text{epochs}} = 4\), and \(\mu = 0.3\) is adopted in all subsequent experiments. 


\subsection{Training Process and Convergence}
\begin{figure}[h]
\centerline{\includegraphics[width=\columnwidth]{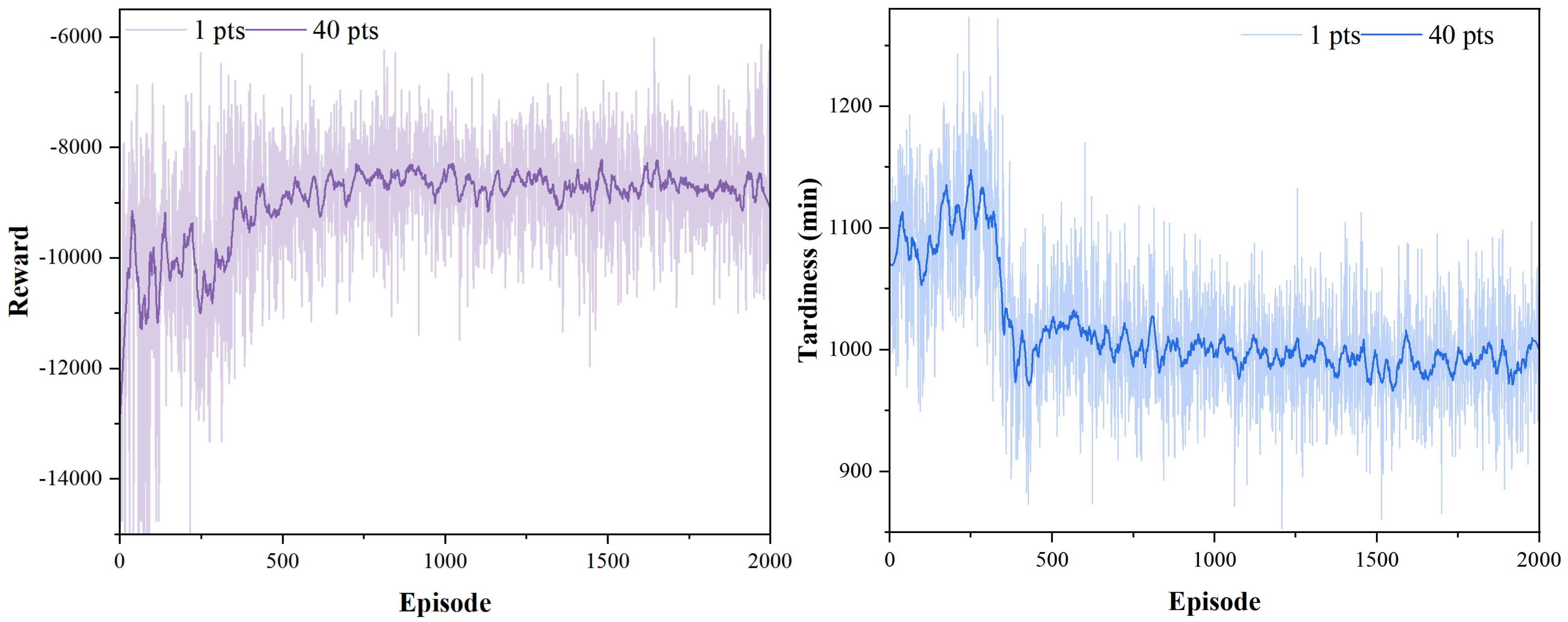}}
\caption{Training trends of reward and tardiness. Solid curves show raw data, and lighter curves indicate 40-point moving averages.}
\label{Fig9}
\end{figure}

Fig.~\ref{Fig9} shows the training curve of SWRL on 30 real-world instances using the Adam optimizer. The cumulative reward increases steadily while the average tardiness decreases, with both metrics stabilizing after approximately 800 episodes. The convergence trend indicates that the spatiotemporal encoding network is able to learn effective job-machine assignment strategies from the sparse reward signals introduced by the dual-layer kitting constraints.

Training variance is high initially due to the large action space. As the sliding-window filtering mechanism focuses attention on kitting-critical nodes, the effective action space shrinks and policy updates stabilize. After convergence, SWRL achieves consistent performance across repeated runs, confirming training stability.

\subsection{Ablation Experiment}
This section conducts ablation studies on 20 real-world instances to assess the contribution of each SWRL module: the sliding-window filtering mechanism, the spatiotemporal graph encoder, and the constrained waiting strategy. Two baseline methods are included: the EDD dispatching rule, and DQN for adaptive rule selection. All end-to-end variants share the same heterogeneous graph backbone~\cite{song2022flexible} and are trained for 2,000 episodes. The module configuration and performance are summarized in Table~\ref{tab04}.

\begin{table}[t]
\caption{Ablation study: module configuration and performance}\label{tab04}
\begin{adjustbox}{center}
\centering
\resizebox{1\columnwidth}{!}{
\begin{threeparttable}
\begin{tabular}{l|ccc|ccc}
\toprule
\multirow{2}{*}{Method}
& \multicolumn{3}{c|}{Module} & \multicolumn{3}{c}{Tardiness} \\
\cmidrule(l){2-4} \cmidrule(l){5-7}
& SW & ST & CW
& ALL $T_{\text{avg}}$ & Train $T$ & Test $T$ \\
\midrule
EDD (dispatching rule)           & -- & -- & -- & 3128.3 & 3409.0 & 2847.5 \\
DQN (rule-selection DRL)         & -- & -- & -- & 2389.7 & 2259.6 & 2519.7 \\
\midrule
HGNN backbone                    & -- & -- & -- & 2233.9 & 2077.9 & 2233.9 \\
$\quad$ + CW                    & -- & -- & $\checkmark$ & 2176.1 & 2115.7 & 2176.1 \\
$\quad$ + ST                    & -- & $\checkmark$ & -- & 1897.1 & 1939.5 & 1897.1 \\
$\quad$ + SW                    & $\checkmark$ & $\checkmark$ & -- & 1773.9 & \textbf{1699.6} & 1848.2 \\
$\quad$ + SWRL                  & $\checkmark$ & $\checkmark$ & $\checkmark$ & \textbf{1689.2} & 1725.2 & \textbf{1653.2} \\
\bottomrule
\end{tabular}
\begin{tablenotes}
\footnotesize
\item Module acronyms: SW = Sliding Window, ST = Spatiotemporal encoder, CW = Constrained Waiting strategy. $\checkmark$ indicates the module is enabled.
\item CPU times (ms) are: EDD 0.875, DQN 5.482, backbone 11.355, +CW 11.454, +ST 12.284, +SW 16.063, SWRL 14.079.

\end{tablenotes}
\end{threeparttable}
}
\end{adjustbox}
\end{table}

Each SWRL module contributes measurable improvements. The HGNN backbone alone achieves $T_{\text{avg}} = 2233.9$, already outperforming both DQN and EDD. Adding the constrained waiting action yields a modest improvement, while the spatiotemporal encoder provides a substantial reduction (-15.1\%), confirming that tracking bottleneck shifts across consecutive decision steps is critical in FAFSP. The sliding-window filtering mechanism further reduces tardiness, and the full SWRL achieves the best overall performance with $T_{\text{avg}} = 1689.2$ on all instances and $T_{\text{avg}} = 1653.2$ on the test set.

Inference time is measured as wall-clock time from event trigger to action selection. All end-to-end variants complete within 11--16 ms, well below typical manufacturing decision intervals (seconds to minutes). Notably, SWRL (14.079 ms) runs faster than SW (16.063 ms) despite including additional modules, because the constrained waiting mechanism reduces the candidate action set and lowers the policy network's computational load. This confirms that the performance gains of SWRL do not come at the cost of inference efficiency.

\begin{figure}[h]
\centerline{\includegraphics[width=1\columnwidth]{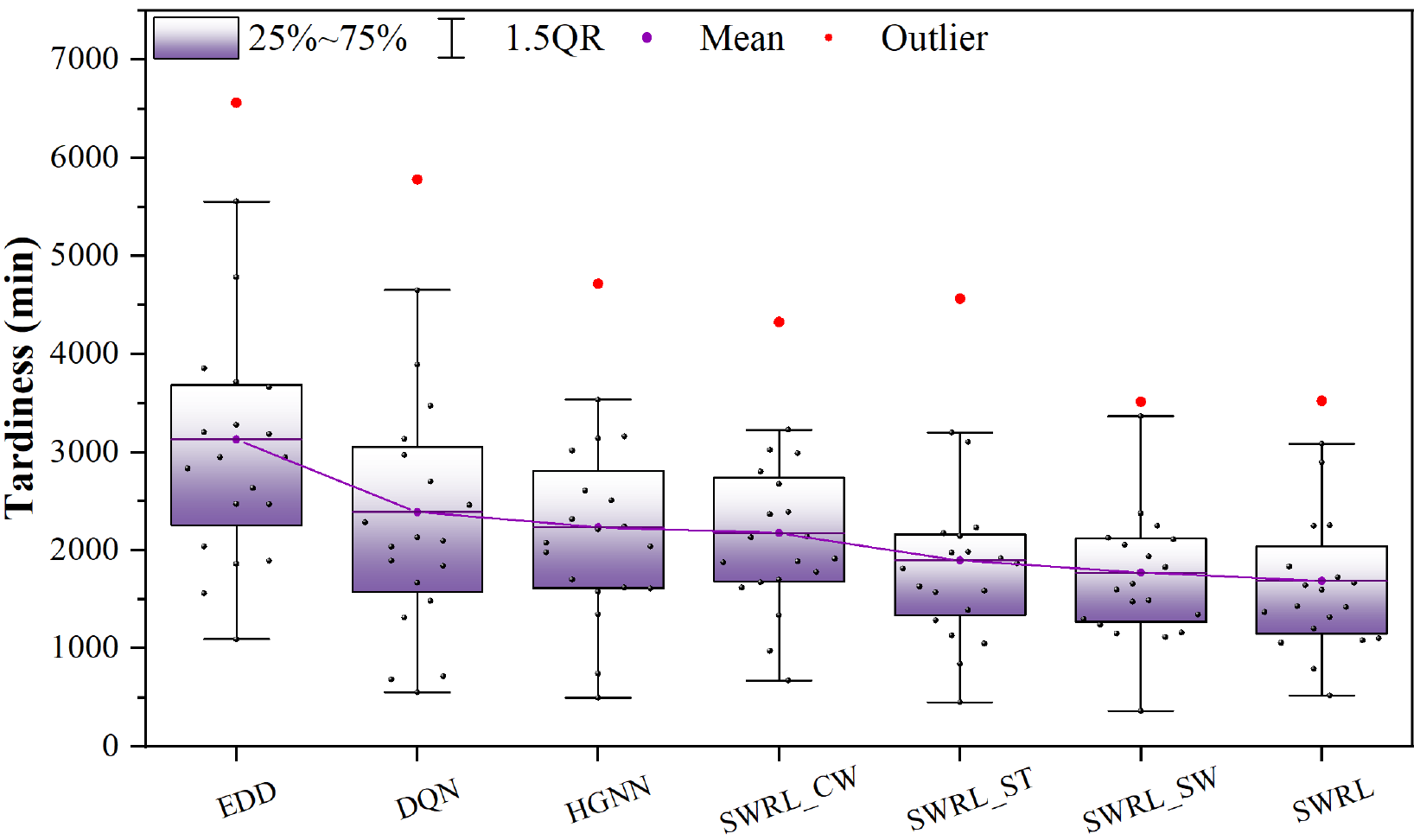}}
\caption{Performance comparison of SWRL and its ablation variants against EDD and DQN in all instances.}
\label{Fig10}
\end{figure}

\begin{figure}[h]
\centerline{\includegraphics[width=0.85\columnwidth]{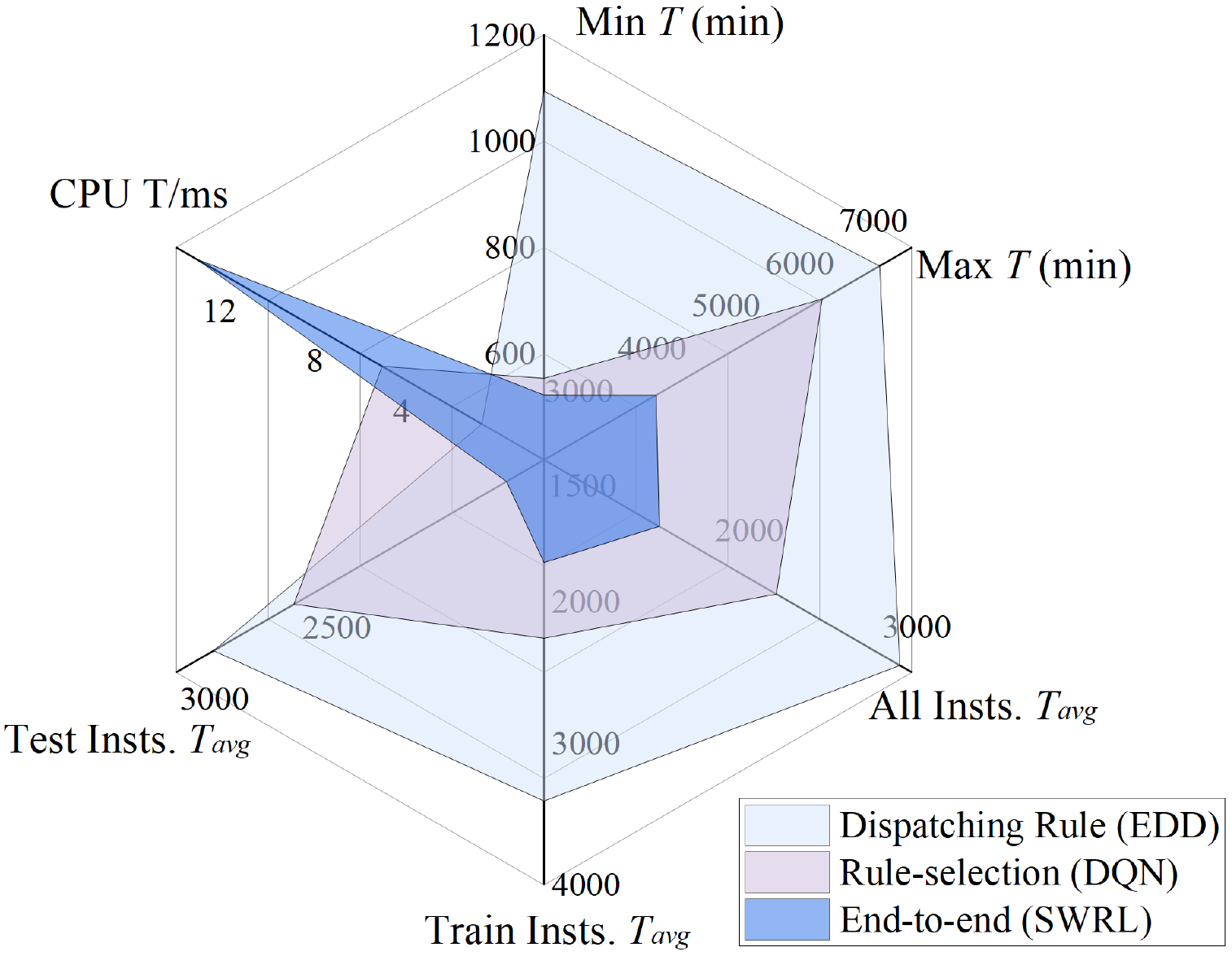}}
\caption{Computation time per decision step across methods. Despite higher graph-processing overhead, SWRL remains within the real-time scheduling requirement.}
\label{Fig11}
\end{figure}

As shown in Fig.~\ref{Fig10}, SWRL consistently outperforms both the rule-based EDD and the DQN-based rule-selection approach across all evaluation metrics. Despite the overhead of graph-structured feature processing, SWRL maintains a practical decision time of 14.079 ms per step (Fig.~\ref{Fig11}), confirming its feasibility for real-time scheduling. 

\subsection{Comparative Experiments}
This section compares SWRL against representative DRL-based scheduling methods, none of which directly address the dynamic FAFSP with multi-product delivery. The rule-selection baselines include Q-learning~\cite{yang2022real}, DQN~\cite{luo2020dynamic}, DDPG~\cite{zhao2022hyperheuristic}, and MLAC-DQN~\cite{qiu2024multi}. The end-to-end baselines include HGNN~\cite{song2022flexible} and DANIEL~\cite{wang2023flexible}, both designed for standard FJSP without assembly kitting constraints. SWRL introduces three domain-specific improvements: (i) a sliding-window filtering mechanism, (ii) a spatiotemporal graph encoder, and (iii) a constrained waiting strategy, which are absent in existing end-to-end FJSP methods. The implementation is developed on the basis of open-source scheduling benchmarks~\cite{reijnen2026job}. The MDP and training configuration follow~\cite{qiu2024multi}, with rule-selection methods trained for 20,000 iterations and end-to-end methods for 2,000 episodes.

\begin{table}[t]
\caption{Comparative results across 20 test instances}\label{tab05}
\begin{adjustbox}{center}
\centering
\resizebox{1\columnwidth}{!}{
\begin{threeparttable}
\begin{tabular}{c|cccc|ccc}
\toprule
\multirow{2}{*}{Inst.}
& \multicolumn{4}{c|}{Rule-selection DRL} & \multicolumn{3}{c}{End-to-end DRL} \\
\cmidrule(l){2-5} \cmidrule(l){6-8}
& Q learing & DQN & DDPG & MLAC-DQN & HGNN & DANIEL & SWRL \\
\midrule
1  & 1674.3 & 1486.5 & 2418.5 & 1678.9 & \textbf{498.2} & 581.0 & 522.0 \\
2  & 1963.1 & 2703.8 & 2663.7 & 2971.8 & 2215.2 & 2050.9 & \textbf{1595.8} \\
3  & 2198.5 & 2971.8 & 2027.4 & \textbf{467.2} & 745.1 & 1566.5 & 1321.0 \\
4  & 3256.5 & 2133.8 & 3247.1 & \textbf{2026.4} & 4717.7 & 3756.6 & 3522.3 \\
5  & 2874.6 & 2037.5 & 2779.6 & 1315.7 & 1580.0 & 1524.3 & \textbf{1204.0} \\
6  & 1467.6 & 1315.7 & 1334.2 & \textbf{1111.7} & 2509.1 & 2045.1 & 2255.6 \\
7  & 1964.7 & \textbf{1669.2} & 2252.1 & \textbf{1669.2} & 3141.7 & 2432.7 & 2897.4 \\
8  & 2848.1 & 2098.5 & 2631.7 & 2098.5 & 1704.0 & 1927.2 & \textbf{1419.8} \\
9  & 5101.4 & 3892.9 & 4908.8 & \textbf{711.7} & 2316.8 & 1878.0 & 1433.4 \\
10 & 3062.0 & 2286.6 & 3550.0 & 1953.3 & 1351.0 & 1425.2 & \textbf{1081.0} \\
11 & 2169.8 & 1897.8 & \textbf{1423.2} & 1897.8 & 3534.1 & 3545.0 & 3084.4 \\
12 & 633.6  & \textbf{553.2}  & 850.1  & \textbf{553.2}  & 3017.2 & 3057.1 & 1644.1 \\
13 & \textbf{1748.3} & 1840.0 & 2543.8 & 1840.0 & 2244.7 & 1980.5 & 2249.4 \\
14 & 4025.3 & 3474.8 & 4478.0 & 2898.1 & 1619.3 & 2053.1 & \textbf{1059.1} \\
15 & 5188.1 & 5780.2 & 6103.6 & \textbf{1450.3} & 1980.8 & 2843.6 & 1725.4 \\
16 & 4347.0 & 4652.2 & 4032.6 & 2985.6 & 2609.1 & 2944.6 & \textbf{1103.7} \\
17 & 794.3  & 685.0  & \textbf{684.1} & 685.0  & 2077.9 & 841.7 & 789.8  \\
18 & 2436.3 & 2461.5 & 2700.7 & 2925.8 & 2042.6 & \textbf{1754.4} & 1834.3 \\
19 & 2671.4 & 3135.7 & 3474.1 & 3135.7 & 3162.4 & 3441.1 & \textbf{1371.0} \\
20 & 1241.0 & \textbf{717.1} & 1086.2 & \textbf{717.1} & 1611.0 & 1709.5 & 1670.7 \\
\midrule
$T_{\text{avg}}$ & 2583.3 & 2389.7 & 2759.5 & 1754.7 & 2233.9 & 2167.9 & \textbf{1689.2} \\
$\sigma$         & 1294.5 & 1326.7 & 1401.2 & 887.4  & 973.1  & 863.5  & \textbf{772.1} \\
W/T/L            & 13/0/7 & 12/0/8 & 13/0/7 & 9/0/11 & 16/0/4 & 16/0/4 & -- \\
\bottomrule
\end{tabular}
\begin{tablenotes}
\footnotesize
\item W/T/L: instances where SWRL achieves a lower (W), equal (T), or higher (L) tardiness than the column method. Other methods are trained for 20,000 episodes; SWRL, HGNN and DANIEL for 2,000 episodes.
\end{tablenotes}
\end{threeparttable}
}
\end{adjustbox}
\end{table}

As shown in Table~\ref{tab05}, Q-learning and DDPG, despite performing well in continuous action spaces, fail to achieve satisfactory results in online FAFSP, with average tardiness of 2583.3 and 2759.5, respectively. Even with DDPG's actor-critic deterministic policy structure, capturing the complex state-to-rule mapping remains challenging. DQN, adept at discrete action spaces, consistently ranks among the top three across most instances. After 20,000 training epochs, MLAC-DQN outperforms others on several cases by decomposing multi-stage rules to expand the solution space and selecting rule combinations across multiple workshops.

In contrast, SWRL achieves the lowest average tardiness (1689.2) with only 2,000 training episodes, while also exhibiting the lowest standard deviation ($\sigma = 772.1$). In pairwise comparisons, SWRL outperforms Q-learning, DQN, DDPG, and HGNN in the majority of instances (13/20, 12/20, 13/20, and 16/20 wins, respectively), achieves a comparable 9/11 split against MLAC-DQN, and also exceeds DANIEL (16/20 wins). Within the end-to-end group, SWRL achieves the best overall performance ($T_{\text{avg}} = 1689.2$), followed by DANIEL ($T_{\text{avg}} = 2167.9$), which outperforms the HGNN backbone ($T_{\text{avg}} = 2233.9$). A Friedman test confirms statistically significant differences across all methods ($p = 0.019$).

\subsection{Robustness Verification}
The sliding-window filtering mechanism is designed to filter irrelevant nodes and maintain focus on kitting-critical operations regardless of the total order volume. This design is expected to confer robustness across different problem scales and resource configurations. To verify this, SWRL is evaluated under four experimental conditions that vary in resource availability (\(m_1 \times m_2\)), number of orders (\(i_{\text{total}}\)), and arrival concentration (\(d\)), as described in Table~\ref{tab06}. Each condition includes 40 instances, with 20 used for training and 20 for testing, totaling 160 instances. SWRL is compared against nine composite dispatching rules commonly used in industrial practice~\cite{BAYKASOGLU2010369}.

\begin{table}[t]
\caption{Robustness comparison across four operating conditions}\label{tab06}
\begin{adjustbox}{center}
\centering
\resizebox{1\columnwidth}{!}{
\begin{threeparttable}
\begin{tabular}{l|cccc}
\toprule
Method & I & II & III & IV \\
& \multicolumn{1}{c}{(0.3,3$\times$6,5)} & \multicolumn{1}{c}{(0.5,5$\times$10,10)} & \multicolumn{1}{c}{(0.7,7$\times$10,15)} & \multicolumn{1}{c}{(0.9,10$\times$15,30)} \\
\midrule
EDD          &  452.0 &  776.2 & 3173.6 & 18004.3 \\
FIFO+SPT     &  426.9 &  785.4 & 2908.7 & 17302.3 \\
FIFO+EET     &  440.8 &  808.6 & 3054.6 & 17852.8 \\
MOPNR+SPT    &  501.3 &  796.7 & \textbf{2899.0} & \textbf{17301.3} \\
MOPNR+EET    &  512.8 &  835.2 & 3090.7 & 17845.2 \\
LWKR+SPT     & \textbf{369.8} & \textbf{750.9} & 2911.5 & 17320.8 \\
LWKR+EET     &  379.2 &  787.0 & 3069.0 & 17860.0 \\
MWKR+SPT     &  486.5 &  770.2 & 2903.8 & 17318.9 \\
MWKR+EET     &  489.2 &  794.1 & 3079.6 & 17866.3 \\
\midrule
Best rule    & 369.8  & 750.9 & 2899.0  & 17301.3 \\
SWRL         & \textbf{341.5}    & \textbf{644.8}    & \textbf{2353.1}    & \textbf{14478.4} \\
Improvement  & +7.7\%          & +14.2\%          & +18.8\%           & +16.3\% \\
\bottomrule
\end{tabular}
\begin{tablenotes}
\footnotesize
\item Each condition contains 40 instances, totaling 160 instances across all conditions.
\end{tablenotes}
\end{threeparttable}
}
\end{adjustbox}
\end{table}

The dispatching rules exhibit strong condition-dependent performance. LWKR+SPT achieves the best rule-level performance in Conditions I and II (small scale), while MOPNR+SPT becomes optimal in Conditions III and IV (larger scale), confirming that the best rule shifts with resource availability and order load. Selecting the appropriate rule for a given operating condition would require expert knowledge of the production environment, and even the best rule lags significantly behind SWRL.

SWRL consistently achieves the lowest tardiness across all four conditions, with improvements over the best dispatching rule ranging from 7.7\% (Condition I) to 18.8\% (Condition III). Notably, the improvement grows with problem scale, indicating that the sliding-window filtering mechanism and spatiotemporal encoding become increasingly valuable as the scheduling complexity increases. This consistent advantage demonstrates that DRL-based end-to-end scheduling adapts to diverse operating environments without manual rule engineering, making it robust for real-world production settings with fluctuating conditions.

\section{Conclusion}
\label{Conclusion}
This paper proposed a sliding-window-based reinforcement learning framework for online scheduling in the dynamic flexible assembly flow shop problem with multi-product delivery. The dual-layer kitting constraints were formalized as a heterogeneous graph-based MDP that captures the tail-product bottleneck dynamics and the resulting sparse reward structure. The SWRL framework integrates three components designed to address the specific challenges of DFAFSP-MPD: a sliding-window filtering mechanism that filters inactive nodes and prioritizes kitting-critical operations under evolving topologies, a spatiotemporal graph encoding network that tracks bottleneck shifts across consecutive decision states, and a dynamic action mapping module with a constrained waiting strategy that handles changing action spaces and sparse feedback.

Experimental results on real-world instances from a home appliance manufacturer demonstrated that SWRL achieves consistent improvements over classical dispatching rules and existing DRL methods across varying resource configurations, order loads, and arrival concentrations. Ablation studies confirmed the contribution of each designed module to the overall performance, and the parameter sensitivity analysis provided guidelines for practical deployment.

Limitations of the current framework include a fixed sliding-window size and the assumption of stationary machine availability, which may limit generalization to varying problem scales or sudden breakdowns. Future work will address these limitations through adaptive window sizing and online fault-adaptation mechanisms, and will extend the framework to incorporate additional constraints such as automated guided vehicle scheduling and energy-aware objectives. Transfer learning across different shop floor configurations and the joint optimization of sliding-window selection and policy learning are also promising directions.

\bibliography{IEEEabrv,Manu}

\begin{thebibliography}{10}
\providecommand{\url}[1]{#1}
\csname url@samestyle\endcsname
\providecommand{\newblock}{\relax}
\providecommand{\bibinfo}[2]{#2}
\providecommand{\BIBentrySTDinterwordspacing}{\spaceskip=0pt\relax}
\providecommand{\BIBentryALTinterwordstretchfactor}{4}
\providecommand{\BIBentryALTinterwordspacing}{\spaceskip=\fontdimen2\font plus
\BIBentryALTinterwordstretchfactor\fontdimen3\font minus \fontdimen4\font\relax}
\providecommand{\BIBforeignlanguage}[2]{{%
\expandafter\ifx\csname l@#1\endcsname\relax
\typeout{** WARNING: IEEEtran.bst: No hyphenation pattern has been}%
\typeout{** loaded for the language `#1'. Using the pattern for}%
\typeout{** the default language instead.}%
\else
\language=\csname l@#1\endcsname
\fi
#2}}
\providecommand{\BIBdecl}{\relax}
\BIBdecl

\bibitem{8821409}
Y.~Fang, C.~Peng, P.~Lou, Z.~Zhou, J.~Hu, and J.~Yan, ``{D}igital-{T}win-{B}ased {J}ob {S}hop {S}cheduling {T}oward {S}mart {M}anufacturing,'' \emph{{IEEE} Trans. Ind. Informat.}, vol.~15, no.~12, pp. 6425--6435, 2019.

\bibitem{chen2022unmanned}
J.~Chen, J.~Sun, and G.~Wang, ``From unmanned systems to autonomous intelligent systems,'' \emph{Engineering}, vol.~12, pp. 16--19, 2022.

\bibitem{Zhao2018Manufacturing}
C.~Zhao, N.~Kang, J.~Li, and J.~A. Horst, ``{P}roduction {C}ontrol to {R}educe {S}tarvation in a {P}artially {F}lexible {P}roduction-{I}nventory {S}ystem,'' \emph{{IEEE} Trans. Autom. Control}, vol.~63, no.~2, pp. 477--491, 2018.

\bibitem{liu2025recent}
T.~Liu, J.~Liu, J.~Qiu, C.~Lai, and Z.~Zhang, ``Recent advances in assembly flow shop scheduling and its extensions to distributed manufacturing and integrated supply chain,'' \emph{Comput. Ind. Eng.}, p. 111711, 2025.

\bibitem{liao2025collaborative}
R.~Liao, J.~Liu, J.~Qiu, and C.~Peng, ``{C}ollaborative optimisation framework for multi-stage flexible assembly shop scheduling with mixed production pattern,'' \emph{Int. J. Prod. Res.}, pp. 1--19, 2025.

\bibitem{Liao2015efficient}
C.-J. Liao, C.-H. Lee, and H.-C. Lee, ``{A}n efficient heuristic for a two-stage assembly scheduling problem with batch setup times to minimize makespan,'' \emph{Comput. Ind. Eng.}, vol.~88, pp. 317--325, 2015.

\bibitem{Wu2021Metaheuristics}
C.-C. Wu, X.~Zhang, A.~Azzouz, W.-L. Shen, S.-R. Cheng, P.-H. Hsu, and W.-C. Lin, ``{M}etaheuristics for two-stage flow-shop assembly problem with a truncation learning function,'' \emph{Eng. Optim.}, vol.~53, no.~5, pp. 843--866, 2021.

\bibitem{li2025dynamic}
Q.-Y. Li, Q.-K. Pan, L.~Wang, L.~Gao, and W.-M. Li, ``Dynamic cascaded flow-shop scheduling using an evolutionary greedy algorithm,'' \emph{{IEEE} Trans. Evol. Comput.}, 2025.

\bibitem{9711566}
G.~Zhang, B.~Liu, L.~Wang, D.~Yu, and K.~Xing, ``{D}istributed {C}o-{E}volutionary {M}emetic {A}lgorithm for {D}istributed {H}ybrid {D}ifferentiation {F}lowshop {S}cheduling {P}roblem,'' \emph{{IEEE} Trans. Evol. Comput.}, vol.~26, no.~5, pp. 1043--1057, 2022.

\bibitem{lei2023large}
K.~Lei, P.~Guo, Y.~Wang, J.~Zhang, X.~Meng, and L.~Qian, ``{L}arge-scale dynamic scheduling for flexible job-shop with random arrivals of new jobs by hierarchical reinforcement learning,'' \emph{{IEEE} Trans. Ind. Informat.}, vol.~20, no.~1, pp. 1007--1018, 2023.

\bibitem{Hatami2015Heuristics}
S.~Hatami, R.~Ruiz, and C.~Andr{\'e}s-Romano, ``{H}euristics and metaheuristics for the distributed assembly permutation flowshop scheduling problem with sequence dependent setup times,'' \emph{Int. J. Prod. Econ.}, vol. 169, pp. 76--88, 2015.

\bibitem{feng2016robust}
X.~Feng, F.~Zheng, and Y.~Xu, ``{R}obust scheduling of a two-stage hybrid flow shop with uncertain interval processing times,'' \emph{Int. J. Prod. Res.}, vol.~54, no.~12, pp. 3706--3717, 2016.

\bibitem{rahmani2016stable}
D.~Rahmani and R.~Ramezanian, ``{A} stable reactive approach in dynamic flexible flow shop scheduling with unexpected disruptions: {A} case study,'' \emph{Comput. Ind. Eng.}, vol.~98, pp. 360--372, 2016.

\bibitem{guo2024multi}
P.~Guo, H.~Shi, Y.~Wang, and J.~Xiong, ``{M}ulti-objective scheduling of cloud-edge cooperation in distributed manufacturing via multi-agent deep reinforcement learning,'' \emph{Int. J. Prod. Res.}, pp. 1--25, 2024.

\bibitem{yang2022real}
S.~Yang, J.~Wang, and Z.~Xu, ``{R}eal-time scheduling for distributed permutation flowshops with dynamic job arrivals using deep reinforcement learning,'' \emph{Adv. Eng. Inf.}, vol.~54, p. 101776, 2022.

\bibitem{li2025real}
Y.~Li, Q.~Wang, X.~Li, L.~Gao, L.~Fu, Y.~Yu, and W.~Zhou, ``{R}eal-time scheduling for flexible job shop with {A}gvs using multiagent reinforcement learning and efficient action decoding,'' \emph{{IEEE} Trans. Syst., Man, Cybern., Syst.}, 2025.

\bibitem{mnih2015human}
V.~Mnih, K.~Kavukcuoglu, D.~Silver, A.~A. Rusu, J.~Veness, M.~G. Bellemare, A.~Graves, M.~Riedmiller, A.~K. Fidjeland, G.~Ostrovski \emph{et~al.}, ``{H}uman-level control through deep reinforcement learning,'' \emph{Nature}, vol. 518, no. 7540, pp. 529--533, 2015.

\bibitem{wang2021review}
L.~Wang, Z.~Pan, and J.~Wang, ``A review of reinforcement learning based intelligent optimization for manufacturing scheduling,'' \emph{Complex Syst. Model. Simul.}, vol.~1, no.~4, pp. 257--270, 2021.

\bibitem{zhang2023counterfactual}
N.~Zhang, Y.~Shen, Y.~Du, L.~Chen, and X.~Zhang, ``{C}ounterfactual-attention multi-agent reinforcement learning for joint condition-based maintenance and production scheduling,'' \emph{J. Manuf. Syst.}, vol.~71, pp. 70--81, 2023.

\bibitem{9673698}
L.~Wang, Z.~Pan, and J.~Wang, ``{A} review of {R}einforcement {L}earning {B}ased {I}ntelligent {O}ptimization for {M}anufacturing {S}cheduling,'' \emph{Complex Syst. Model. Simul.}, vol.~1, no.~4, pp. 257--270, 2021.

\bibitem{luo2020dynamic}
S.~Luo, ``{D}ynamic scheduling for flexible job shop with new job insertions by deep reinforcement learning,'' \emph{Appl. Soft Comput.}, vol.~91, p. 106208, 2020.

\bibitem{yang2023real}
S.~Yang, J.~Wang, L.~Xin, and Z.~Xu, ``{R}eal-time and concurrent optimization of scheduling and reconfiguration for dynamic reconfigurable flow shop using deep reinforcement learning,'' \emph{CIRP J. Manuf. Sci. Technol.}, vol.~40, pp. 243--252, 2023.

\bibitem{wang2022multi}
H.~Wang, J.~Cheng, C.~Liu, Y.~Zhang, S.~Hu, and L.~Chen, ``{M}ulti-objective reinforcement learning framework for dynamic flexible job shop scheduling problem with uncertain events,'' \emph{Appl. Soft Comput.}, vol. 131, p. 109717, 2022.

\bibitem{wang2022independent}
M.~Wang, J.~Zhang, P.~Zhang, L.~Cui, and G.~Zhang, ``{I}ndependent double {D}qn-based multi-agent reinforcement learning approach for online two-stage hybrid flow shop scheduling with batch machines,'' \emph{J. Manuf. Syst.}, vol.~65, pp. 694--708, 2022.

\bibitem{qiu2024multi}
J.~Qiu, J.~Liu, Z.~Li, and X.~Lai, ``{A} multi-level action coupling reinforcement learning approach for online two-stage flexible assembly flow shop scheduling,'' \emph{J. Manuf. Syst.}, vol.~76, pp. 351--370, 2024.

\bibitem{luo2021real}
S.~Luo, L.~Zhang, and Y.~Fan, ``{R}eal-time scheduling for dynamic partial-no-wait multiobjective flexible job shop by deep reinforcement learning,'' \emph{{IEEE} Trans. Autom. Sci. Eng.}, vol.~19, no.~4, pp. 3020--3038, 2021.

\bibitem{wang2021adaptive}
H.~Wang, B.~R. Sarker, J.~Li, and J.~Li, ``{A}daptive scheduling for assembly job shop with uncertain assembly times based on dual {Q}-learning,'' \emph{Int. J. Prod. Res.}, vol.~59, no.~19, pp. 5867--5883, 2021.

\bibitem{liu2022graph}
Z.~Liu, Y.~Wang, X.~Liang, Y.~Ma, Y.~Feng, G.~Cheng, and Z.~Liu, ``{A} graph neural networks-based deep {Q}-learning approach for job shop scheduling problems in traffic management,'' \emph{Inf. Sci.}, vol. 607, pp. 1211--1223, 2022.

\bibitem{li2025graph}
Y.~Li, Q.~Liu, C.~Zhang, X.~Li, and L.~Gao, ``Graph-based dual-agent deep reinforcement learning for dynamic human--machine hybrid reconfiguration manufacturing scheduling,'' \emph{{IEEE} Trans. Syst., Man, Cybern., Syst.}, 2025.

\bibitem{liu2023dynamic}
C.-L. Liu and T.-H. Huang, ``Dynamic job-shop scheduling problems using graph neural network and deep reinforcement learning,'' \emph{{IEEE} Trans. Syst., Man, Cybern., Syst.}, vol.~53, no.~11, pp. 6836--6848, 2023.

\bibitem{song2022flexible}
W.~Song, X.~Chen, Q.~Li, and Z.~Cao, ``{F}lexible job-shop scheduling via graph neural network and deep reinforcement learning,'' \emph{{IEEE} Trans. Ind. Informat.}, vol.~19, no.~2, pp. 1600--1610, 2022.

\bibitem{zhang2022dynamic}
Y.~Zhang, H.~Zhu, D.~Tang, T.~Zhou, and Y.~Gui, ``{D}ynamic job shop scheduling based on deep reinforcement learning for multi-agent manufacturing systems,'' \emph{Rob. Comput. Integr. Manuf.}, vol.~78, p. 102412, 2022.

\bibitem{lei2022multi}
K.~Lei, P.~Guo, W.~Zhao, Y.~Wang, L.~Qian, X.~Meng, and L.~Tang, ``{A} multi-action deep reinforcement learning framework for flexible {J}ob-shop scheduling problem,'' \emph{Expert Syst. Appl.}, vol. 205, p. 117796, 2022.

\bibitem{park2026deep}
J.~Park, Y.~Su, and F.~Ju, ``Deep reinforcement learning with dynamic graph pruning for scalable flexible job shop scheduling,'' \emph{IISE Trans.}, no. just-accepted, pp. 1--31, 2026.

\bibitem{challenge1}
D.~Johnson, G.~Chen, and Y.~Lu, ``Multi-agent continuous decision-making for the continuous dynamic flexible job shop scheduling problem,'' \emph{{IEEE} Trans. Autom. Sci. Eng.}, 2026.

\bibitem{basir2018bi}
S.~A. Basir, M.~M. Mazdeh, and M.~Namakshenas, ``{B}i-level genetic algorithms for a two-stage assembly flow-shop scheduling problem with batch delivery system,'' \emph{Comput. Ind. Eng.}, vol. 126, pp. 217--231, 2018.

\bibitem{allahverdi2016minimizing}
A.~Allahverdi, A.~Aydilek, and H.~Aydilek, ``{M}inimizing the number of tardy jobs on a two-stage assembly flowshop,'' \emph{J. Ind. Prod. Eng.}, vol.~33, no.~6, pp. 391--403, 2016.

\bibitem{wang2023flexible}
R.~Wang, G.~Wang, J.~Sun, F.~Deng, and J.~Chen, ``Flexible job shop scheduling via dual attention network-based reinforcement learning,'' \emph{IEEE Trans. Neural Netw. Learn. Syst.}, vol.~35, no.~3, pp. 3091--3102, 2023.

\bibitem{zhao2022hyperheuristic}
F.~Zhao, S.~Di, and L.~Wang, ``{A} hyperheuristic with {Q}-learning for the multiobjective energy-efficient distributed blocking flow shop scheduling problem,'' \emph{{IEEE} Trans. Cybern.}, vol.~53, no.~5, pp. 3337--3350, 2022.

\bibitem{reijnen2026job}
R.~Reijnen, I.~G. Smit, H.~Zhang, Y.~Wu, Z.~Bukhsh, and Y.~Zhang, ``Job shop scheduling benchmark: Environments and instances for learning and non-learning methods,'' \emph{Ann. Math. Artif. Intell.}, pp. 1--31, 2026.

\bibitem{BAYKASOGLU2010369}
A.~Baykaso{\u g}lu and L.~{\"O}zbak{\i}r, ``{A}nalyzing the effect of dispatching rules on the scheduling performance through grammar based flexible scheduling system,'' \emph{Int. J. Prod. Econ.}, vol. 124, no.~2, pp. 369--381, 2010.

\end{thebibliography}
\bibliographystyle{IEEEtran}

\end{document}